\documentclass[sigconf]{acmart}

\AtBeginDocument{%
  }

\setcopyright{cc}
\setcctype{by}

\copyrightyear{2026}
\acmYear{2026}

\acmConference[CIKM '26]
{Proceedings of the 35th ACM International Conference on Information and Knowledge Management}
{November 7--11, 2026}
{Rome, Italy}

\acmBooktitle{Proceedings of the 35th ACM International Conference on Information and Knowledge Management
(CIKM '26), November 7--11, 2026, Rome, Italy}

\acmISBN{979-8-4007-2539-5/2026/11}

\acmDOI{10.1145/3799682.3840737}

\usepackage{microtype}
\usepackage{graphicx}
\usepackage{subcaption}
\usepackage{booktabs}

\usepackage{amsmath}

\usepackage{amssymb}
\usepackage{mathtools}
\usepackage{amsthm}
\usepackage{url}
\usepackage[utf8]{inputenc} 
\usepackage[T1]{fontenc}    
\usepackage{nicefrac}  
\usepackage{xcolor}  
\usepackage{xspace}
\usepackage{enumitem}
\usepackage{multirow}
\usepackage{wrapfig}
\usepackage{colortbl}
\usepackage{lipsum}
\usepackage{tabularx}
\usepackage{array}
\usepackage{algorithm}
\usepackage{algorithmic}
\usepackage{xurl}
\usepackage{tabularray}
\usepackage{marvosym}
\usepackage{makecell}
\usepackage{hyperref}

\usepackage{stfloats}
\usepackage{caption}
\usepackage[capitalize,noabbrev]{cleveref}

\theoremstyle{plain}

\theoremstyle{definition}

\theoremstyle{remark}

\definecolor{darkgreen}{rgb}{0.0, 0.5, 0.0}

\newcommand{\blue}[1]{\textcolor{blue!85!black}{#1}}
\newcommand{\lightblue}[1]{\textcolor{cyan!70!black}{#1}}
\newcommand{\khaki}[1]{\textcolor{yellow!65!black}{#1}}
\newcommand{\pink}[1]{\textcolor{pink!60!magenta}{#1}}
\newcommand{\red}[1]{\textcolor{red!85!black}{#1}}

\newcommand{\Ours}[0]{Time-R1\xspace}

\begin{document}

\title{Time Series Forecasting via Reasoning: \\A Slow-Thinking Approach with Reinforcement Fine-Tuned LLMs}




\author{Yitong Zhou}
\affiliation{%
  \institution{State Key Laboratory of Cognitive Intelligence, University of Science and Technology of China}
  \city{Hefei}
  \state{Anhui Province}
  \country{China}
}
\email{yitong.zhou@mail.ustc.edu.cn}

\author{Yucong Luo}
\affiliation{%
  \institution{State Key Laboratory of Cognitive Intelligence, University of Science and Technology of China}
  \city{Hefei}
  \state{Anhui Province}
  \country{China}
}
\email{prime666@mail.ustc.edu.cn}

\author{Mingyue Cheng}
\authornote{Corresponding author.}
\affiliation{%
  \institution{State Key Laboratory of Cognitive Intelligence, University of Science and Technology of China}
  \city{Hefei}
  \state{Anhui Province}
  \country{China}
}
\email{mycheng@ustc.edu.cn}

\author{Qi Liu}
\affiliation{%
  \institution{State Key Laboratory of Cognitive Intelligence, University of Science and Technology of China}
  \city{Hefei}
  \state{Anhui Province}
  \country{China}
}
\email{qiliuql@ustc.edu.cn}

\author{Jiahao Wang}
\affiliation{%
  \institution{State Key Laboratory of Cognitive Intelligence, University of Science and Technology of China}
  \city{Hefei}
  \state{Anhui Province}
  \country{China}
}
\email{jiahao.wang@mail.ustc.edu.cn}

\author{Daoyu Wang}
\affiliation{%
  \institution{State Key Laboratory of Cognitive Intelligence, University of Science and Technology of China}
  \city{Hefei}
  \state{Anhui Province}
  \country{China}
}
\email{wdy030428@mail.ustc.edu.cn}

\author{Enhong Chen}
\affiliation{%
  \institution{State Key Laboratory of Cognitive Intelligence, University of Science and Technology of China}
  \city{Hefei}
  \state{Anhui Province}
  \country{China}
}
\email{cheneh@ustc.edu.cn}

\renewcommand{\shortauthors}{Yitong Zhou et al.}



\begin{abstract}
To advance time series forecasting, a wide range of methods—from statistical techniques to deep learning architectures—have been proposed to improve prediction accuracy; however, most still follow a \textit{fast thinking} paradigm that maps historical patterns to future values with limited explicit reasoning or interpretable intermediate steps, undermining adaptability to evolving temporal dynamics.
Although slow-thinking LLMs exhibit strong multi-step reasoning, prompt-only solutions are constrained by cost, privacy, and insufficient domain alignment, motivating the training of LLMs to acquire \textit{slow thinking} time-series reasoning skills.
In this paper, we propose \Ours, a reinforcement fine-tuning framework designed to enhance the multi-step reasoning ability of LLMs for time series forecasting. 
Training starts with supervised fine-tuning for warmup adaptation, followed by reinforcement learning that improves the model’s generalization ability via reward-driven reasoning optimization.
Specifically, we introduce a fine-grained multi-objective reward to provide training signals tailored to time series forecasting, and GRIP (group-based relative importance for policy optimization) to leverage non-uniform sampling for encouraging the exploration of effective reasoning paths.
Experiments show that \Ours achieves competitive forecasting performance across diverse datasets
\footnote{Code is at \url{https://github.com/ustc-time-series/Time-R1}}.
\end{abstract}

\maketitle

\vspace{-0.1in}
\section{Introduction}

Time series forecasting (TSF) plays a key role in data-driven decision-making across critical domains, including financial market analysis~\cite{sako2022neural}, energy demand planning~\cite{kotzur2018time}, and traffic flow management~\cite{li2015trend}. Over the years, a spectrum of approaches—from classical statistical models to machine learning and deep learning—has been developed to improve forecasting accuracy.

Classical statistical methods such as ARIMA \citep{zhang2003time}, ETS \citep{hyndman2008forecasting}, and Theta \citep{assimakopoulos2000theta} have long been used to predict future data by leveraging the statistical properties of single samples. Machine learning methods such as XGBoost\citep{chen2016xgboost} and LightGBM\citep{ke2017lightgbm} remain highly effective due to their interpretability and ability to model nonlinear relationships. Recent advancements in deep learning techniques have made them particularly effective at modeling complex temporal dependencies and non-stationary data in time series forecasting. Methodological analysis covers pioneering architectures such as sequence dependency modeling of RNNs \citep{hewamalage2021recurrent,salinas2020deepar}, TCNs \citep{hewage2020temporal} and transformer-based models \citep{ijcai2023p759}, which improve generalization through shared representations across multiple time series.

\begin{figure*}[t] 
    \centering
    \includegraphics[width=0.96\textwidth]{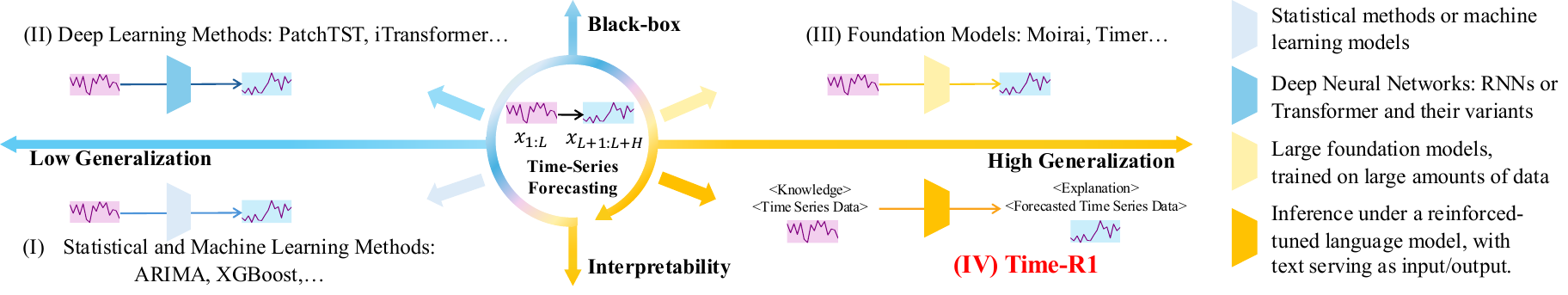} 
    \vspace{-0.1in}
    \caption{A taxonomy of TSF paradigms from statistical/ML models to deep learning and foundation models, organized by increasing generalization and model transparency. \Ours represents a reasoning-oriented paradigm that leverages time-series knowledge and explanation generation to support forecasting.} 
    \vspace{-0.1in}
    \label{fig:7_comparison} 
\end{figure*}

Although the specific techniques vary, most existing TSF methods follow a similar "fast thinking" paradigm~\citep{wang2024tabletime,wang2025can,liu2025evaluating}. Specifically focusing on single-step prediction accuracy, these methods typically employ sequential models to encode historical values and use single-pass direct prediction to directly map past observations to future values. Although effective in benchmarks, their underlying logic is largely based on pattern recognition and trend prediction, lacking an explicit reasoning process. However, in real-world scenarios, time series often reflect more complex temporal logic, which should not merely be 'fitted'—they should be understood and reasoned.

To address this issue, a growing body of work explores how large language models (LLMs) can support time series forecasting. Specifically, recent methods use LLMs to analyze temporal dynamics and learn informative representations, which in turn improve TSF models \citep{chang2023llm4ts,liu2025calf,jin2023time}. These approaches leverage LLMs to incorporate contextual information, such as textual metadata \citep{gruver2023large}. This additional context often improves cross-domain generalization \citep{liu2024time, dooley2023forecastpfn} and enables explanations that support forecasting decisions \citep{tan2024language}.

However, despite their potential, current LLM-based TSF methods face three key limitations:
\textit{First}, a partial misalignment of time series domain knowledge, and limited reasoning capabilities. General linguistic knowledge in LLMs often mismatches the temporal patterns and causal mechanisms required for time series tasks, leading to suboptimal performance \citep{zhou2023one}.
\textit{Second}, a lack of generalization from experiential learning. While effective in fitting historical patterns, they struggle with understanding dynamics or adapting to new, unseen scenarios, which limits their out-of-distribution performance.
\textit{Third}, absence of progressive reasoning. These models map history to future directly without detecting regime changes or performing step-by-step inference, resembling fast (not deliberate) thinking for time series.
These issues lead to a central question: \textbf{Can we improve time series forecasting performance by training LLMs to acquire time series reasoning capabilities?} 

Motivated by this question, we propose \Ours, a novel LLM-based time series forecasting framework that trains large language models to acquire slow-thinking reasoning capabilities for future-trend forecasting. 
\Ours adopts an LLM as the reasoning backbone and follows a curriculum-style RFT optimization: it first warm-starts reasoning and formatting with synthetic step-by-step temporal analyses, and then progressively improves generalization through reward-driven policy optimization.
\textit{First}, we begin with warm-up supervised fine-tuning. The model is fine-tuned for supervised pattern alignment, learning both effective reasoning patterns and accurate output formatting using synthetic reasoning trajectories that demonstrate step-by-step temporal analysis. \textit{Second}, the model is refined through reinforcement learning for generalization, using fine-grained, multi-objective rewards specifically designed for forecasting tasks, improving temporal coherence and multi-horizon accuracy. Notably, we propose GRIP (group-based relative importance for policy optimization), which optimizes LLM reasoning paths in TSF through a non-uniform sampling strategy and adaptive weighting. Extensive experiments are conducted on real-world datasets, showing that \Ours effectively enhances forecast performance through the slow thinking paradigm. Our main contributions are as follows:

\begin{itemize}[leftmargin=10pt, itemsep=1pt]
  \item We introduce time series reasoning by training LLMs to adopt a slow-thinking paradigm that reasons explicitly over the temporal patterns for final forecasting.
  \item We design a RFT framework that enhances the reasoning ability of LLMs. We introduce a fine-grained reward specifically for TSF, along with a novel sampling strategy for RL optimization.
  \item Extensive experiments demonstrate the effectiveness of \Ours, showing that it enhances LLM reasoning and further improves generalization and explainability via deliberate slow thinking.
\end{itemize}

\vspace{-0.1in}
\section{Related Work}
\subsection{Time Series Forecasting}
Time series forecasting has evolved from classical models like ARIMA, effective under ideal conditions \citep{box1968some,zhang2003time}, to modern deep learning approaches. 
Machine learning methods \citep{chen2016xgboost,ke2017lightgbm} remain robust due to their interpretability and ability to model nonlinear relationships.  The advent of deep learning introduced sequence-to-sequence models such as Recurrent Neural Networks, which captured temporal dynamics well \citep{hewamalage2021recurrent,salinas2020deepar}. However, RNNs face limitations like restricted receptive fields and error accumulation \citep{salinas2020deepar}. Advanced architectures incorporating self-attention and convolutional networks have since been developed to capture long-range dependencies \citep{lai2018modeling,li2019enhancing}. Concurrently, integrating traditional techniques like trend-seasonal decomposition into neural networks has improved performance \citep{wen2020fast}. 
Additionally, slice-based methods have shown promise in long-term forecasting by segmenting time series for better accuracy \citep{nie2022time,zhang2023crossformer}. These advancements blend classical principles with deep learning to tackle the challenges of TSF.

\vspace{-0.06in}

\subsection{LLM-based Time Series Forecasting}
In recent years, large language models (LLMs) have attracted attention for their ability to understand and generate human-like text, now extending into time series analysis \citep{zhang2024large,jiang2024empowering,jin2024position}. The application of LLMs in this field primarily follows two approaches: fine-tuning and prompt-based zero-shot learning. Fine-tuning involves further training pre-trained LLMs on specific time series data to improve performance \citep{chang2023llm4ts,chang2024align,liu2025calf,jin2023time}, though it requires significant labeled data. Conversely, prompt-based zero-shot methods utilize the model's existing knowledge through carefully designed prompts, avoiding task-specific training \citep{gruver2023large,liu2024lstprompt}.
More recently, LLM-based time series studies have moved beyond direct forecasting toward reasoning-oriented temporal understanding, exploring context-aware forecasting, time-series question answering, anomaly interpretation, and slow-thinking forecasting \citep{wang2025itformer,kong2025time, ni2026streasoner}. Along this direction, more complex frameworks further organize temporal reasoning around perception, extrapolation, decision-making, and pattern-aware question answering \citep{guan2025timeomni,lu2026patra, tao2026cast,tao2026memcast}.
These advances suggest that effective LLM-based time series modeling requires explicit alignment with temporal patterns and reasoning processes, motivating us to train LLMs with slow-thinking forecasting ability rather than relying solely on prompting or generic adaptation.

\begin{figure*}[t] 
    \centering
    \includegraphics[width=0.94\textwidth]{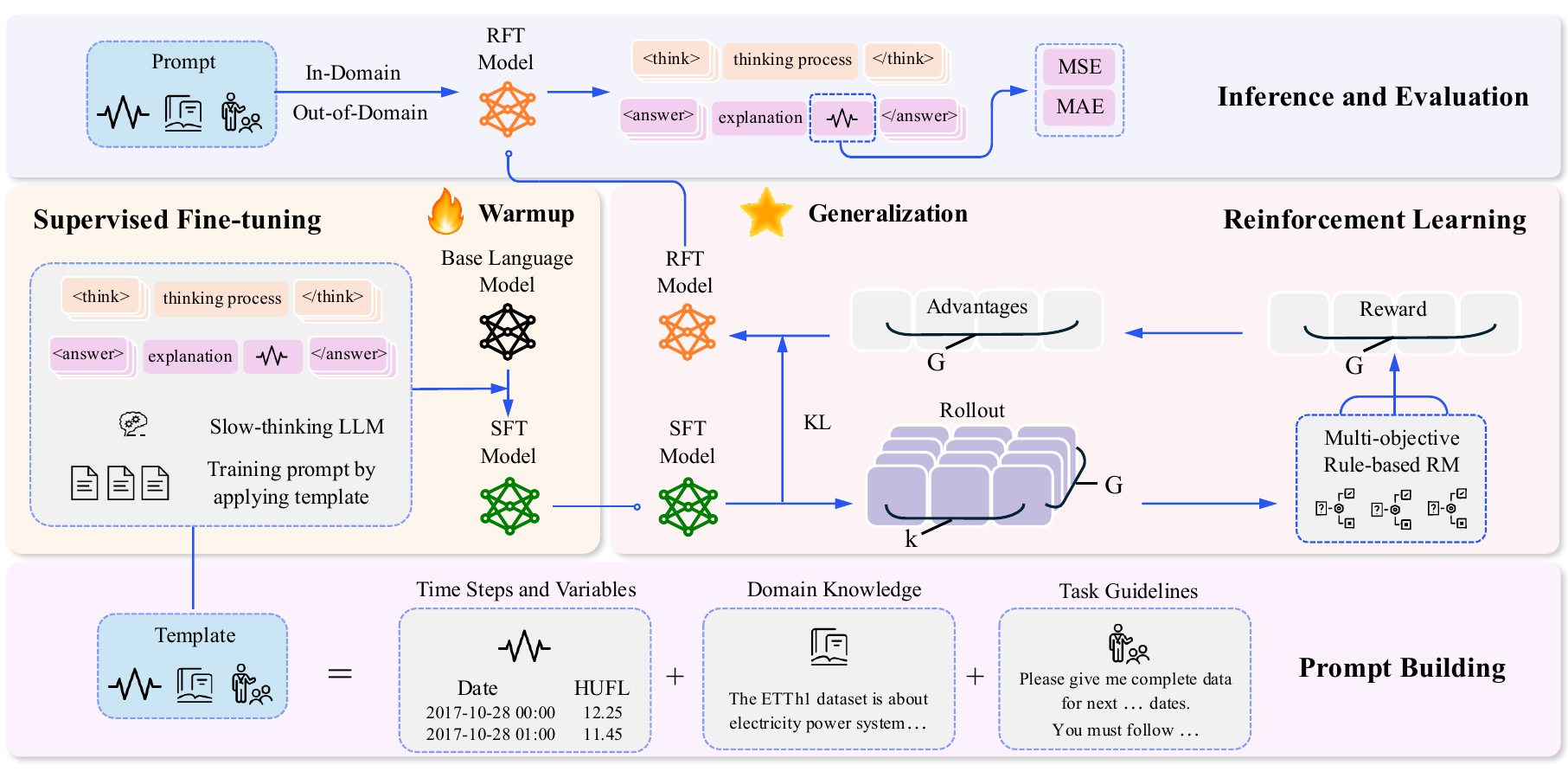} 
    \vspace{-0.15in}
    \caption{A diagram illustrating the three steps of \Ours: (1) building a training template with domain context, time steps, and variables, (2) collecting long-CoT data to train a supervised policy, and (3) optimizing the policy via reinforcement learning with group-based relative importance for policy optimization (GRIP) to enhance TSF reasoning capability.} 
    \vspace{-0.06in}
    \label{fig:1_framework} 
\end{figure*}

\vspace{-0.06in}
\subsection{LLMs and Reinforcement Learning}
Reinforcement Learning (RL) \citep{kaelbling1996reinforcement} allows an agent to learn decision-making through interactions with its environment, aiming to maximize cumulative rewards. RLHF introduced RL to LLMs via human feedback \citep{ouyang2022training, kaufmann2023survey}, initially training a reward model on human preferences and then using it for tuning the policy LLM, often with Proximal Policy Optimization.
To simplify this, methods like Direct Preference Optimization \citep{rafailov2023direct} and SimPO \citep{meng2024simpo} have been proposed, offering computational efficiency but suffering from off-policy issues \citep{pang2024iterative}. 
Another approach, GRPO \citep{shao2024deepseekmath}, avoids a critic model by estimating baselines from group scores. Following this line, DAPO \citep{yu2026dapo} improves group-based RL with decoupled clipping and dynamic sampling, while GSPO \citep{zheng2025group} performs policy optimization at the sequence level for more stable training. 
Recently, there has been growing interest in combining CoT reasoning with RL to improve reasoning quality and self-refinement in LLMs~\citep{chu2025sft, chen2025sft, song2025r1, li2024getting}. Despite these advances, applying RL to enhance LLM-driven reasoning for time series forecasting tasks remains underexplored in practical, real-world forecasting settings today.

\vspace{-0.06in}
\section{The Proposed \Ours}

\subsection{Problem Definition}

Let $\mathbb{D} = \{(X^i,y^i)\}_{i=1}^n$ be a temporal dataset, where each $X^i \in \mathbb{R}^{t \times m}$ is a multivariate time series with $t$ steps and $m$ channels, and $y^i \in \mathbb{R}^{h \times d}$ contains $d$-dimensional targets over $h$ future steps. The forecasting task learns a mapping $f_\theta: \mathbb{R}^{t \times m} \rightarrow \mathbb{R}^{h \times d}$ capturing temporal dependencies in $\mathbb{D}$. Under \Ours, the forecasting procedure using prompt template $P$ is: $T^i = \text{LLM}_\phi(P, X^i)$, where $T^i$ is the LLM's textual output, and $\hat{y}^i = g(T^i)$ parses it into numerical predictions $\hat{y}^i \in \mathbb{R}^{h \times d}$. Here, $g: \mathcal{T}\rightarrow \mathbb{R}^{h \times d}$ denotes a deterministic parsing function that extracts $h\!\cdot\! d$ real-valued numbers from $T^i$ and reshapes them into an $h \times d$ matrix.

\subsection{\Ours Overview}

\Ours consists of a two-stage RFT framework for LLM-based time series forecasting, built upon a structured training template that standardizes input representations and encodes task-specific knowledge. In the first stage, we perform warm-up SFT using synthetic chain-of-thought trajectories to teach the model effective temporal analysis and accurate output formatting. These trajectories are generated under strict guidelines and refined iteratively to align with ground-truth forecasts. The second stage further improves the model via RL, guided by a fine-grained, multi-objective reward function tailored for time series forecasting. To optimize reasoning paths during RL, we introduce GRIP (Group-based Relative Importance for Policy Optimization), a novel strategy that leverages non-uniform sampling and adaptive weighting to balance accuracy, logical consistency, and temporal coherence. An overview of the framework is provided in Figure \ref{fig:1_framework}.


\begin{table}[ht]
\centering

\caption{Training prompt template for Time-R1. Contents enclosed in \{\} will be replaced with specific information, including dataset description, channel information, and historical time-series data.} 
\vspace{-0.08in}
\label{tab:training_template}
\renewcommand{\arraystretch}{0.9}
\begin{tabular}{@{} p{0.5\dimexpr\textwidth-4\tabcolsep} @{}}
\toprule
Here is the \blue{\{Channel Information\}} data of the ETTh1 dataset. The ETTh1 dataset is \lightblue{\{Dataset Description\}}. I will now give you data for the past 96 recorded dates, and please help me forecast the data for the next 96 recorded dates accurately. The data is as follows: \khaki{\{Historical Time Series Data\}}. Please give me the complete data for the next 96 recorded dates in chronological order. Remember to give me the complete data without omission. You must first conduct reasoning inside \pink{<think>}...\pink{</think>}. When you have the final answer, you can output the answer inside \red{<answer>}...\red{</answer>}. \\
\bottomrule
\end{tabular}
\vspace{-0.08in}
\end{table}

\vspace{-0.12in}
\subsection{Training Template}

Our training template standardizes inputs and encodes task-specific knowledge through five carefully designed components: (1) \textit{Task Definition} establishing objectives and problem scope; (2) \textit{Dataset Description} specifying temporal characteristics and application scenarios; (3) \textit{Channel Information} which delineates the types and semantic meanings of input signals; (4) \textit{Testing Data} providing timestamps and historical series; and (5) \textit{Format Instruction} defining output templates. 
By organizing these elements into a unified prompt structure, the template interleaves domain knowledge with explicit structural constraints, enabling the model to better understand the forecasting context while reducing inference-time formatting ambiguities and response inconsistencies (see Table \ref{tab:training_template}).

\subsection{SFT for Warmup Adaptation} 
\label{SFTdata}

To mitigate readability degradation and slow convergence caused by direct reinforcement learning on LLMs, we first perform a warmup stage via SFT. This step stabilizes training, ensures proper output formatting, and equips the model with reasoning capabilities.

Our SFT data construction involves three steps. First, we leverage DeepSeek-R1 to generate time-series predictions on the training set by feeding historical time series data with strict formatting guidelines. We then select the optimal prediction for each sample based on MSE. Next, to derive reasoning aligned with ground-truth labels, we inject both the true prediction value and the high-quality CoT generated in the previous step into DeepSeek-R1 as prompts, guiding it to synthesize a revised CoT that logically culminates in the correct prediction. Finally, we concatenate the refined CoT with the true prediction value, demarcating the final answer using `<answer>` tags to create structured SFT data.

After constructing the training data, we perform single-epoch fine-tuning. This warm-up SFT phase prepares the model for reinforcement learning by ensuring stable learning dynamics and accurate output formatting. It also enables the model to internalize reasoning patterns from synthetic trajectories, laying the foundation for more deliberate and coherent decision-making in subsequent reinforcement learning stages.

\subsection{RL for Effective Reasoning}


After warmup SFT, we further fine-tune the LLM with RL to improve reasoning generalization and enable slow-thinking for time series forecasting. We next introduce the reward design and the RL algorithm GRIP.

\subsubsection{Reward Design}
\label{sec:Reward Modeling}
To effectively apply RL for optimizing the proposed slow-thinking time series forecasting, we introduce several fine-grained and multi-objective reward functions specifically designed to enhance forecasting performance and slow thinking behavior. We define the total reward as a weighted sum of the following components.

\paragraph{Format Rewards.}  
To ensure syntactic validity and completeness of the generated reasoning paths, we define two reward components to enforce both structural integrity and output completeness:

\textit{Format Reward:} A binary penalty is imposed if the output does not follow the required structured format (e.g., missing or malformed \texttt{<answer>} tags):
{
\setlength{\abovedisplayskip}{5pt}
\setlength{\belowdisplayskip}{5pt}
\begin{equation}
\small
\gamma_{\text{Format}} = 
\begin{cases}
0 & \text{if valid } \texttt{<think>}, \texttt{<answer>} \text{ tags} \\
-1 & \text{otherwise}
\end{cases}
\end{equation}
}

\textit{Length Reward:} To encourage full time point sequence generation and accelerate convergence, we provide positive feedback based on how close the generated sequence length is to the ground truth:
{\setlength{\abovedisplayskip}{5pt}
\setlength{\belowdisplayskip}{5pt}
\begin{equation}
\gamma_{Length} = 0.1 \cdot \min \left( 1, \frac{L_a}{L_g} \right),
\end{equation}
}
where $L_a$ and $L_g$ denote the predicted-sequence length and the ground-truth-sequence length.

\begin{figure*}[t] 
    \centering
    \includegraphics[width=0.96\textwidth, keepaspectratio]{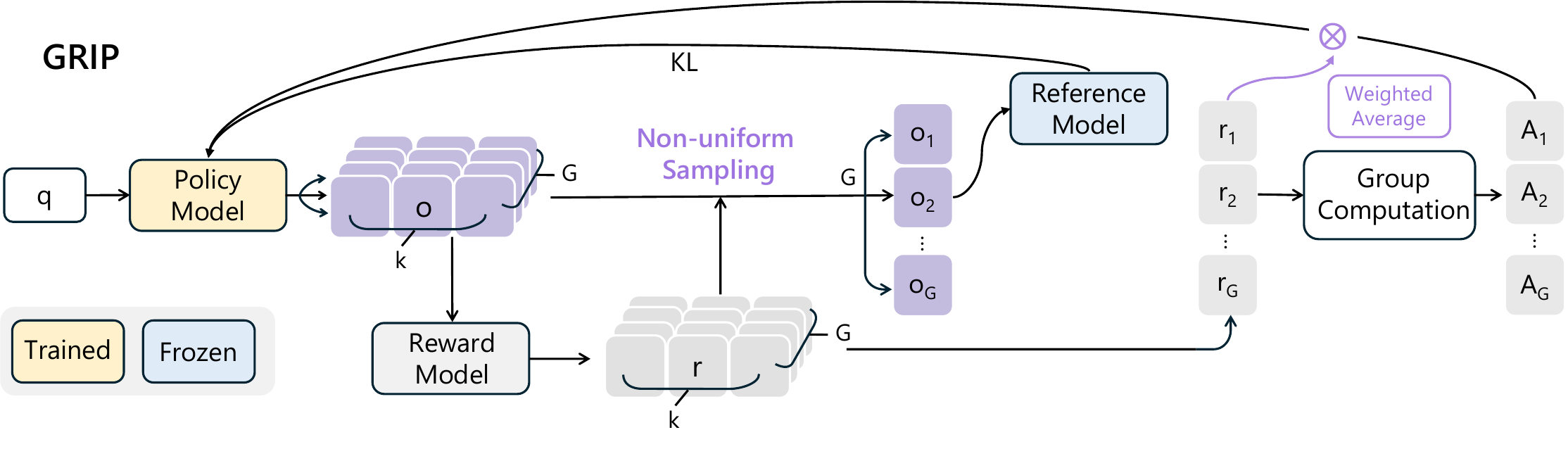} 

    \caption{Overview of Group-based Relative Importance for Policy Optimization (GRIP). } 
    \label{fig:2_grip} 

\end{figure*}

\paragraph{Accuracy Rewards.}
We define two accuracy-based reward components to assess numerical precision, encouraging accurate value prediction and modeling of dynamics:

\textit{MSE Reward:} We compute the mean squared error between normalized prediction and target sequences and map it into a bounded reward signal using a sigmoid transformation:
{\setlength{\abovedisplayskip}{5pt}
\setlength{\belowdisplayskip}{5pt}
\begin{equation}
\gamma_{\text{MSE}} = \left(1 - \frac{1}{1 + e^{-2 \cdot \text{MSE}}}\right) \cdot 2.
\end{equation}
}


\textit{Seasonal--Trend Decomposition Reward:} We decompose both predicted and ground-truth sequences into seasonal $s$ and trend $t$ components, and compute  MSE:
{\setlength{\abovedisplayskip}{5pt}
\setlength{\belowdisplayskip}{5pt}
\begin{align}
\gamma_{\text{Seasonal}} &= \frac{1}{n} \sum_{i=1}^{n} \left(s_i^{\text{true}} - s_i^{\text{pred}}\right)^2
\end{align}
}
{\setlength{\abovedisplayskip}{5pt}
\setlength{\belowdisplayskip}{5pt}
\begin{align}
\gamma_{\text{Trend}}    &= \frac{1}{n} \sum_{i=1}^{n} \left(t_i^{\text{true}} - t_i^{\text{pred}}\right)^2.
\end{align}
}
These two terms first measure the discrepancies in seasonal fluctuations and long-term trends, respectively, and are further mapped into the reward space using a sigmoid-like transformation, thereby providing bounded and smoothly scaled rewards for preserving temporal structures during optimization.

\paragraph{Structural Similarity Reward.}
We evaluate structural similarity by matching predicted and ground-truth extrema, ensuring change-point capture and interpretable patterns, with correct matches receiving credit:
\begin{equation}
\gamma_{\text{CP}} = \left(\frac{N_{\text{cmax}}}{N_{\text{gmax}}} \cdot 0.2 \right) + \left(\frac{N_{\text{cmin}}}{N_{\text{gmin}}} \cdot 0.2\right),
	\setlength{\abovedisplayskip}{2pt}
	\setlength{\belowdisplayskip}{2pt}
\end{equation}
where $N_{\text{cmax}}$ and $N_{\text{cmin}}$ respectively represent the counts of correctly identified local maxima and minima within a tolerance window, $N_{\text{gmax}}$ and $N_{\text{gmin}}$ are the total ground-truth extrema counts.

\subsubsection{Reinforcement Learning Algorithm: GRIP}
We introduce GRIP in Figure~\ref{fig:2_grip}, a general RL optimization method designed to optimize entire trajectories for LLM time series forecasting reasoners. The GRIP objective function, formalized in Equation \ref{eq:grip_obj}, combines a non-uniform sampling strategy with adaptive trajectory weighting within a policy gradient framework. In the following sections, we elaborate on its components:(1) GRIP formalization; (2) non-uniform sampling strategy to balance exploration and exploitation; and (3) an adaptive weighting scheme that enhances gradient signals from high-quality reasoning paths.

\paragraph{Formalization of the GRIP Objective.}

The GRIP objective integrates the two key design components into a unified policy gradient framework. We first define the probability ratio $\rho_t(\theta)$ between the current and old policies as:
\begin{equation}
\setlength{\abovedisplayskip}{2pt}
\setlength{\belowdisplayskip}{2pt}
    \rho_t(\theta) = \frac{\pi_\theta(o_{i,t}|q, o_{i,<t})}{\pi_{\theta_{\text{old}}}(o_{i,t}|q, o_{i,<t})},
    \label{eq:ratio_def}
\end{equation}
the full objective is then formalized in Equation \ref{eq:grip_obj}:

\begin{equation}
\small 
\setlength{\abovedisplayskip}{2pt}
\setlength{\belowdisplayskip}{2pt}
\begin{aligned}
\mathcal{J}_{\text{GRIP}}(\theta) &= \mathbb{E}_{\mathcal{D}} \Bigg[ \sum_{i=1}^{G} \frac{w_i^U}{|o_i|} \sum_{t=1}^{|o_i|} \bigg( \min \Big( \rho_t(\theta) A_i, \\
& \hspace{-5mm}  \text{clip}(\rho_t(\theta), 1-\epsilon, 1+\epsilon) A_i \Big) - \beta \mathbb{D}_{KL} [\pi_\theta || \pi_{\text{ref}}] \bigg) \Bigg],
\end{aligned}
\label{eq:grip_obj}
\end{equation}

where $\mathcal{D}$ represents the sampling process: $q \sim P(Q)$, candidate generation $\{o_j\}_{j=1}^{k \cdot G} \sim \pi_{\theta_{\text{old}}}$, and the selection of trajectories $\{o_i\}_{i=1}^G \sim \text{Sample}(\{o_j\}; R)$. 
In this equation, $\epsilon$ and $\beta$ are hyperparameters, and $\pi_{\text{ref}}$ is the reference model, initialized as the pre-trained model before reinforcement learning begins. The hyperparameter $k$ controls the size of the rollout space, while $G$ is referred to as the group size. $\mathbb{D}_{KL}$ represents the KL divergence, incorporated as a regularization term. Finally, $A_i$ is the advantage computed using a group of rewards $\{r_1, r_2, \dots, r_G\}$ corresponding to the completion trajectories within each group:
\begin{equation}
    A_i = \frac{r_i - \text{mean}(\{r_1, r_2, \dots, r_G\})}{\text{std}(\{r_1, r_2, \dots, r_G\})},
\end{equation}
where $w_i^U$ is the adaptive trajectory weight. This objective balances exploration and exploitation while mitigating gradient dilution.

\begin{table*}[ht]\centering
    \caption{Performance comparison of \Ours and baseline models with best values in bold and second-best underlined. MSE $\downarrow$ is the evaluation metric. DeepSeek-R1 denotes zero-shot using our template. MAE results are reported in Appendix Table~\ref{tab:main_results_mae}.}
    \label{tab:main_results_mse}
    \renewcommand{\arraystretch}{1.2}
    \resizebox{0.98\textwidth}{!}{
    \large
    \begin{tabular}{ll|ccccccccc}
        \toprule
         & \textbf{Methods}    & \textbf{ETTh1}  & \textbf{ETTh2}   & \textbf{ETTm1}   & \textbf{ETTm2}  & \textbf{Exchange} & \textbf{AQWan} & \textbf{AQShunyi} & \textbf{Wind} & \textbf{NASDAQ} \\
        \hline
        \multirow{10}{*}{Traditional} 
        & AutoFormer & 7.3876& 13.9167& 15.2816& 7.6855& 0.0012  & 19628.2867 & 23576.0196      & 1673.8570 & \underline{0.0008}\\
        
        & PatchTST   & 9.3001& 10.9735& 16.3864& 5.8375& 0.0009  & \textbf{12436.8256}& 16693.3076      & 2024.8256 & \textbf{0.0007}   \\
        
        & DLinear    & 7.6954& 10.4067& 13.9395& 7.9100& 0.0014  & 20997.7228 & 20952.4161      & 1619.4311 & \textbf{0.0007}    \\
        
        & iTransformer      & 7.5048& 10.0161& \textbf{12.7511}     & 5.7713& 0.0010  & 13482.5746 & 18219.6612      & 1591.6404  & \underline{0.0008}\\

        & TimeXer    & 8.5213& 11.4268& 14.0023& 5.7325    & 0.0009  & 14397.1884 & 16491.3209   & 1684.9856 & \textbf{0.0007}    \\

        & TimeMixer&     \underline{6.0124}&     \underline{8.8157}&     13.2158&     5.7129&     0.0009&     14128.4175&     16645.6821    & \underline{1380.5264}&     \textbf{0.0007} \\
        
        & WPMixer    & 6.1543    & 8.9326    & 13.3087    & 5.7842    & 0.0009&     13205.8932&     \underline{16220.4375}    & 1402.8173&     \textbf{0.0007} \\

        & Moment & 6.4934 & 9.5210 & 13.7712 & 6.1231 & \underline{0.0008} & 13431.7716 & 17518.0725 & 1490.9685 & \underline{0.0008} \\
        
        & TimesFM-2.5 & 9.9855 & 9.9425 & 36.4792 & 9.4884 & 0.0009 & 13729.7518 & 19037.4867 & 1712.3315 & \textbf{0.0007} \\
        
        & Chronos-2 & 9.4377 & 13.2946 & 33.3549 & 8.7010 & 0.0009 & 13446.7502 & 16635.3240 & 2146.0611 & 0.0010 \\
        
        \hline
        \multirow{4}{*}{LLM-based} & CrossTimeNet &8.3125 & 11.6789 & 16.3475 & 6.7924 & 0.0011 & 15120.0853 & 18042.1278 & 1931.2672 & 0.0012  \\        
        & GPT4TS &  6.9928&  9.7971&  15.8238&  5.7014&  0.0009&  13546.1725&  16839.0718&  1790.3269&  0.0010\\
        & TimeLLM &  6.8780&  9.9814&  15.8845&  \underline{5.6695}&  0.0010&  13427.4982&  16665.2379&  1575.8937&  0.0011\\
        & DeepSeek-R1 & 6.7098 & 11.3845 & 14.8561 & 7.0063 & 0.0026 & 29653.1218 & 30780.9011 & 4047.1201 & 0.0021\\
        \hline
        \cellcolor{lightgray!30}Ours & \cellcolor{lightgray!30}Time-R1    & \cellcolor{lightgray!30}\textbf{5.8752}    & \cellcolor{lightgray!30}\textbf{8.7093}      & \cellcolor{lightgray!30}\underline{13.1034}& \cellcolor{lightgray!30}\textbf{5.6673}& \cellcolor{lightgray!30}\textbf{0.0007}& \cellcolor{lightgray!30}\underline{13033.1820}  & \cellcolor{lightgray!30}\textbf{16150.5556}    & \cellcolor{lightgray!30}\textbf{1353.9381}& \cellcolor{lightgray!30}\textbf{0.0007} \\
        \bottomrule
    \end{tabular}
}
\end{table*}

\paragraph{Non-uniform Sampling Strategy.}
\label{sec:sampling}

To bridge the gap between reasoning and forecasting in time series modeling, recent RL methods like GRPO~\cite{shao2024deepseekmath} have shown promise. However, they often suffer from an exploration-exploitation imbalance. 
In the context of TSF, this limitation is critical: the vast, continuous search space of temporal dynamics makes uniform sampling prone to missing sparse, high-quality reasoning paths, often causing models to stagnate in local minima of simple pattern memorization.

To address this, GRIP introduces a non-uniform sampling strategy that generates $k \cdot G$ candidate trajectories $\{o_j\}_{j=1}^{k \cdot G}$ from policy $\pi_{\theta_{\text{old}}}$ (where $k$ scales exploration and $G$ is group size), then selects $G$ elite trajectories via reward-weighted sampling $\text{Sample}(\{o_j\}; R(o_j))$. These are replicated to form the update set $\{o_i\}_{i=1}^G$, maintaining GRPO's update scale while emphasizing high-reward regions. 
From a policy-gradient perspective, this procedure can be interpreted as an importance-sampled correction that increases the contribution of more informative trajectories without changing the overall group-based optimization form. In this way, GRIP combines a broader rollout space with a compact update set, allowing the model to explore diverse reasoning paths while maintaining computational efficiency. To further generalize this mechanism, GRIP supports two sampling strategies:

(1) \textit{Local Random Sampling}: For each input question $q$, we first generate $k$ candidate trajectories $\{o_j\}_{j=1}^k$ by independently sampling from the old policy $\pi_{\theta_{\text{old}}}$. The trajectory with the highest reward $o^* = \arg\max_{1 \leq j \leq k} R(o_j)$ is selected as the elite sample. This process is repeated $G$ times to construct the final set $\{o_i\}_{i=1}^G$. This strategy emphasizes deterministic exploitation of the top-performing sample at each iteration while maintaining computational efficiency.  

(2) \textit{Cluster-based Random Sampling}: For each $q$, we generate $k \cdot G$ candidate trajectories $\{o_j\}_{j=1}^{k \cdot G}$. These trajectories are clustered based on their rewards (e.g., reward-binning or K-means clustering), and $G$ trajectories are randomly sampled across clusters to ensure diversity in the final update set. This method balances exploration and exploitation by preserving low-reward but potentially informative samples while still prioritizing high-reward paths.

\paragraph{Adaptive Weighting for Gradient Enhancement.}
\label{sec:Adaptive Weighting}
\label{Adaptive Weighting}

In time-series forecasting, rewards are continuous and non-binary, where many trajectories can be partially correct. Under uniform aggregation, the policy may overly reinforce safe but low-entropy behaviors, such as copying recent observations or producing mean-line forecasts, because these trajectories can obtain moderately acceptable rewards despite lacking real predictive insight. To provide more discriminative optimization signals, GRIP assigns trajectory-specific weights based on their relative quality within the selected group:
{\setlength{\abovedisplayskip}{2pt}
\setlength{\belowdisplayskip}{2pt}
\begin{equation}
w_i^U =
\frac{[\hat{x}_{q,o_i}-\bar{x}_q]_+ + \lambda}
{\sum_{j=1}^{G}\left([\hat{x}_{q,o_j}-\bar{x}_q]_+ + \lambda\right)},
\end{equation}}
where $\hat{x}_{q,o_i}$ denotes a trajectory reward $R(o_i)$, $\bar{x}_q=\frac{1}{G}\sum_{j=1}^{G}\hat{x}_{q,o_j}$ is the group-level average score, $[\cdot]_+$ denotes positive clipping, and $\lambda$ is a small smoothing constant. By assigning larger weights to higher-reward trajectories, GRIP amplifies their gradient contributions, thereby encouraging stronger token-level probability updates for tokens associated with high-quality forecasting behaviors. This helps induce higher policy entropy, which implicitly promotes exploration in reinforcement learning. This entropy-preserving behavior expands the exploration space, reduces the risk of premature convergence to suboptimal patterns, and helps the model discover better forecasting trajectories. In this way, GRIP can better adapt to the graded reward landscape of time-series forecasting while still encouraging stable and effective policy improvement.

\vspace{-0.06in}
\section{Experiments}
\vspace{0.04in}
\subsection{Experimental Setup}

\paragraph{Datasets and Evaluation Metrics.}

To ensure comprehensive evaluation across diverse scenarios, we conduct experiments on nine datasets from multiple domains with distinct temporal characteristics (Table~\ref{tab:data_statistics}). These include ETT \citep{zhou2021informer} for 2016--2018 electricity load records, Exchange \citep{lai2018modeling} for 1990--2016 foreign exchange rates, Wind \citep{lai2018modeling} for 2020--2021 wind measurements, AQ \citep{zhang2017cautionary} for four-year air quality records, and NASDAQ \citep{feng2019temporal} for stock market series including prices and trading volumes.
We report MSE between predictions and ground truth under a 96-step forecasting horizon for all datasets, except NASDAQ, which uses a 36-step horizon.


\paragraph{Baselines.}

Our representative baselines include various competitive methods:
Autoformer \citep{wu2021autoformer}, PatchTST \citep{nie2022time}, DLinear \citep{zeng2023transformers}, iTransformer \citep{liu2023itransformer}, TimeXer \citep{wang2024timexer}, TimeMixer \citep{wang2024timemixer}, 
WPMixer \citep{murad2025wpmixer}, Moment \citep{goswami2024moment}, TimesFM-2.5 \citep{das2024decoder} and Chronos-2 \citep{ansari2025chronos}. For recent LLM-based approaches, we include CrossTimeNet \citep{crossTimeNet}, GPT4TS \citep{zhou2023one}, TimeLLM \citep{jin2023time}, and DeepSeek-R1 \citep{deepseekR1} for zero-shot inference.

\paragraph{Implementation Details.}

Tables~\ref{tab:main_results_mse} and~\ref{tab:main_results_mae} report MSE and MAE on each dataset's original scale, with all results averaged over three runs using different random seeds. Baselines follow their official or commonly adopted experimental configurations, with predictions inverse-transformed before evaluation when normalization is used. For \Ours, we keep the model-facing time-series values in the original numerical scale as both inputs and outputs, allowing the LLM to access physically meaningful magnitudes during reasoning.

For \Ours, we use Qwen2.5-7B-Instruct as the backbone. In SFT, we train on 3000 synthetic samples with a learning rate of 5e-5 for one epoch. In RL, we implement GRIP using the VeRL framework \citep{verl} with vLLM for generation, $\epsilon=0.2$ and $\beta=0.04$; group size $G=16$, $k=3$. The batch size is 16, learning rate is 1e-6, policy temperature is 1, and max completion length is 8000. Both stages are run on a 4-GPU A800 cluster. For DeepSeek-R1, we apply its prompt directly to time series prediction without training.


\begin{table}[t]\centering
    \caption{Statistics of the experimental datasets.}
    \renewcommand{\arraystretch}{1.2}
    \vspace{-0.1in}
    \label{tab:data_statistics}
    \resizebox{0.48\textwidth}{!}{
    \large
    \begin{tabular}{cccccccccc}
        \toprule
        Dataset & Domain & Timestamps & Features &  Frequency \\ 
        \hline
        ETTh1\&ETTh2 & Electricity & 17,420 & 7 &  1 hour \\ 
        ETTm1\&ETTm2 & Electricity & 69,680 & 7 &   15 mins \\
        AQWan\&AQShunyi & Environment & 35,064 & 11 & 1 hour  \\ 
        Exchange & Economy & 7,588 & 8 &  1 day  \\ 
        Wind & Energy & 48,673 & 7 &   15 mins \\ 
        NASDAQ & Stock & 1,244 & 5 &  1 day \\ 
        \bottomrule
    \end{tabular}
    }
    \vspace{-0.2in}
\end{table}

\begin{figure*}[htp]
    \centering
    \begin{subfigure}[b]{0.23\textwidth}
        \centering
        \includegraphics[width=\textwidth]{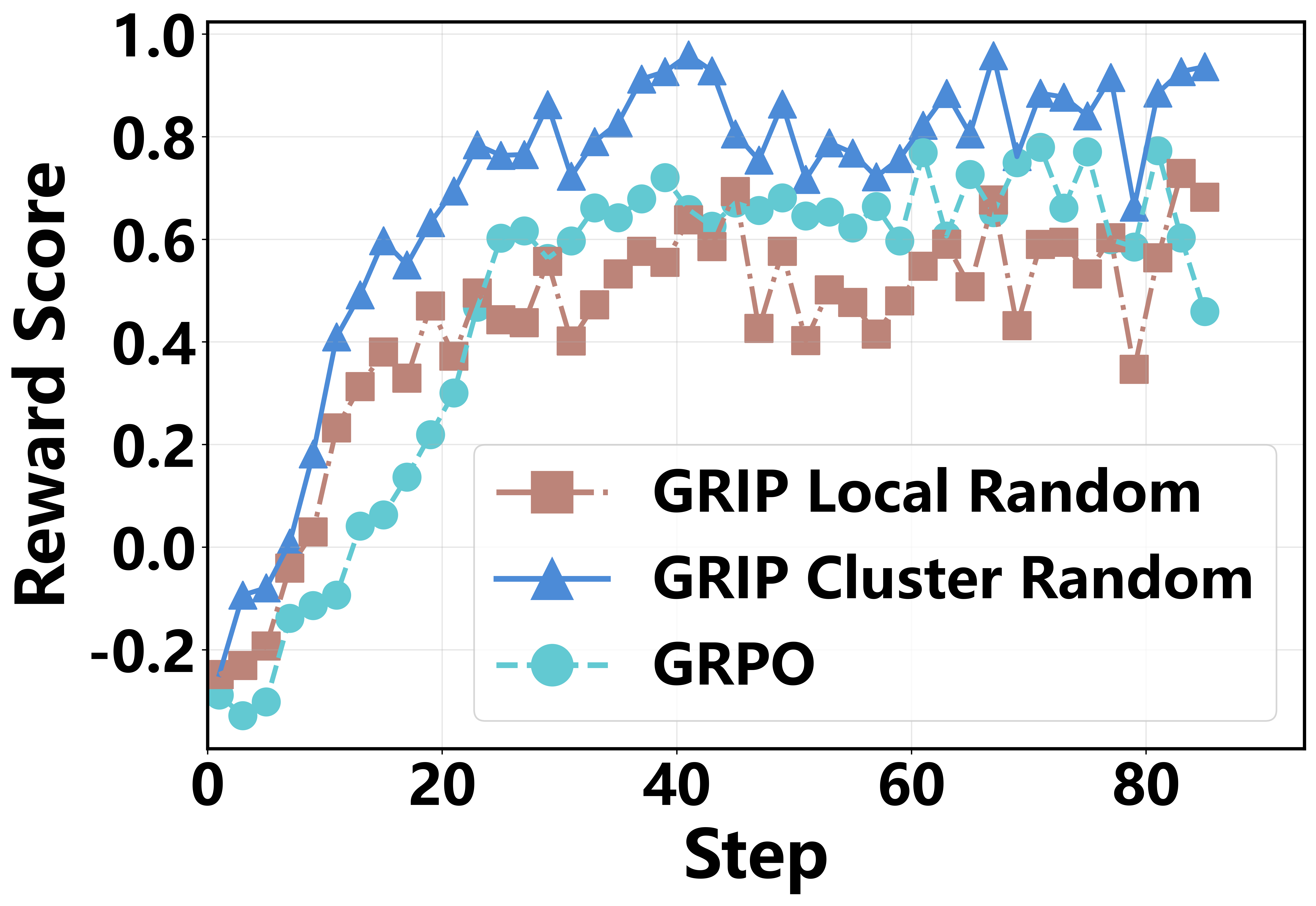}
        \caption{GRIP vs. GRPO}
        \label{fig:compare_a}
    \end{subfigure}
    \hfill
    \begin{subfigure}[b]{0.23\textwidth}
        \centering
        \includegraphics[width=\textwidth]{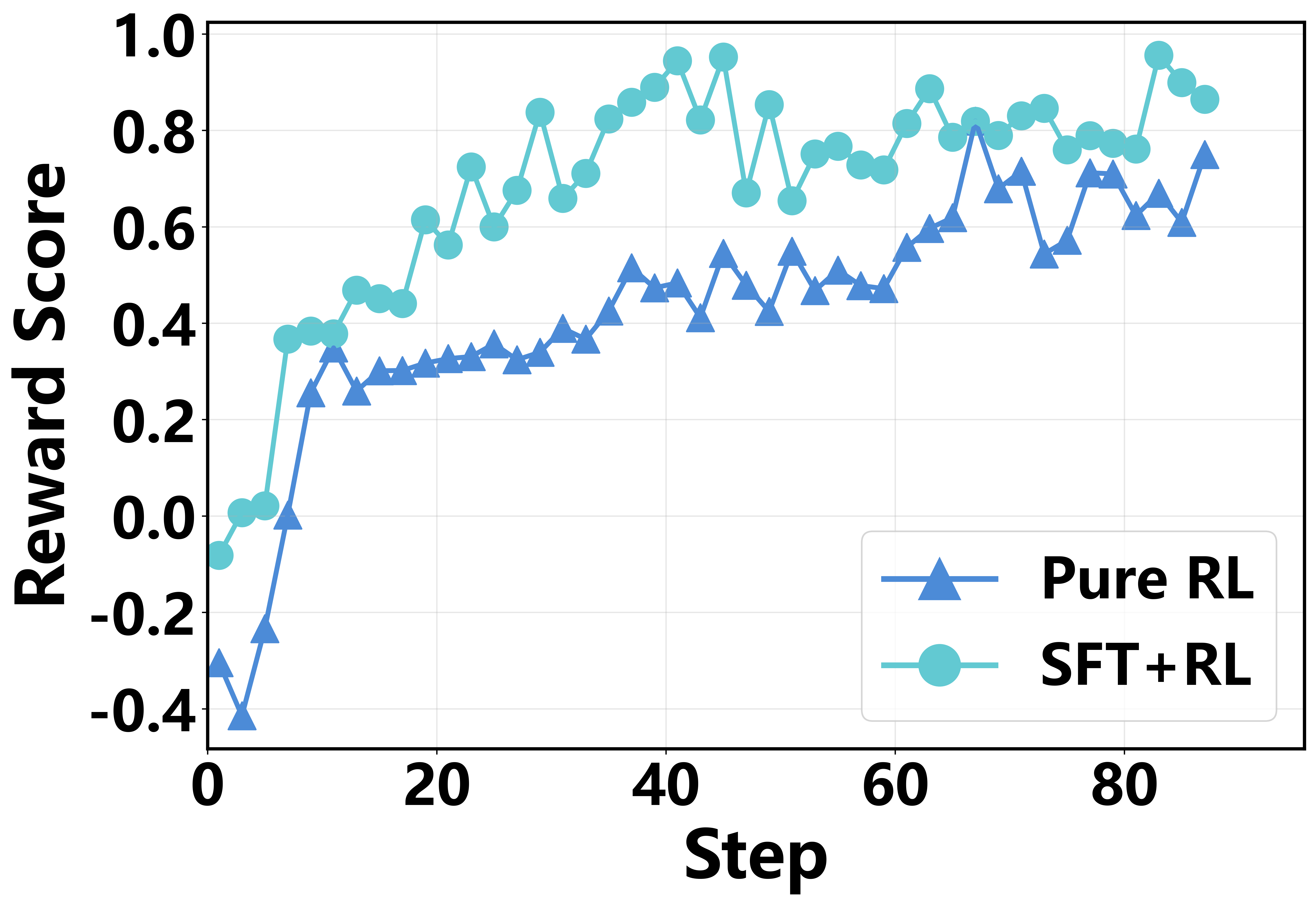}
        \caption{RL vs. SFT + RL}
        \label{fig:compare_b}
    \end{subfigure}
    \hfill
    \begin{subfigure}[b]{0.23\textwidth}
        \centering
        \includegraphics[width=\textwidth]{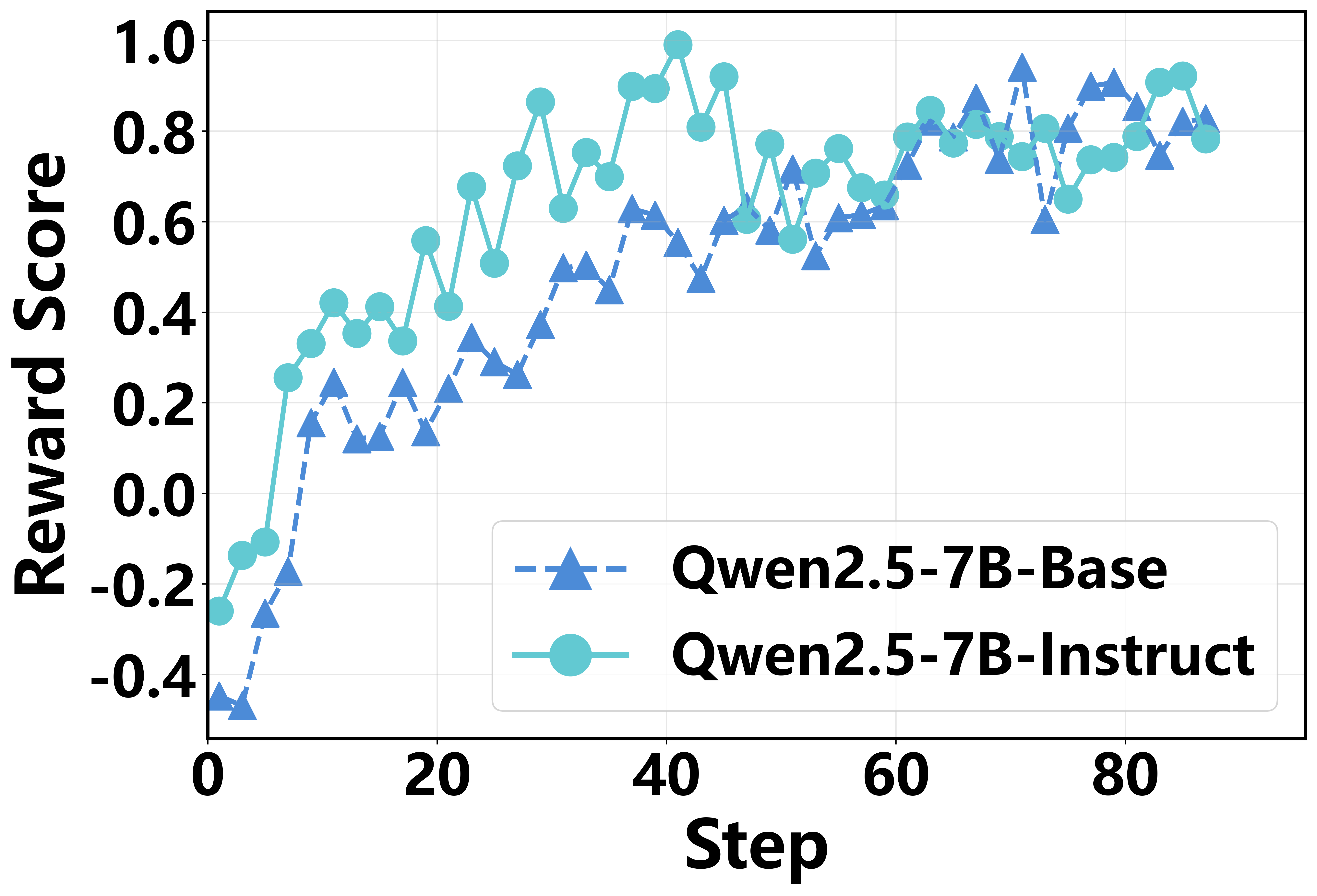}
        \caption{Base vs. Instruct}
        \label{fig:compare_c}
    \end{subfigure}
    \hfill
    \begin{subfigure}[b]{0.23\textwidth}
        \centering
        \includegraphics[width=\textwidth]{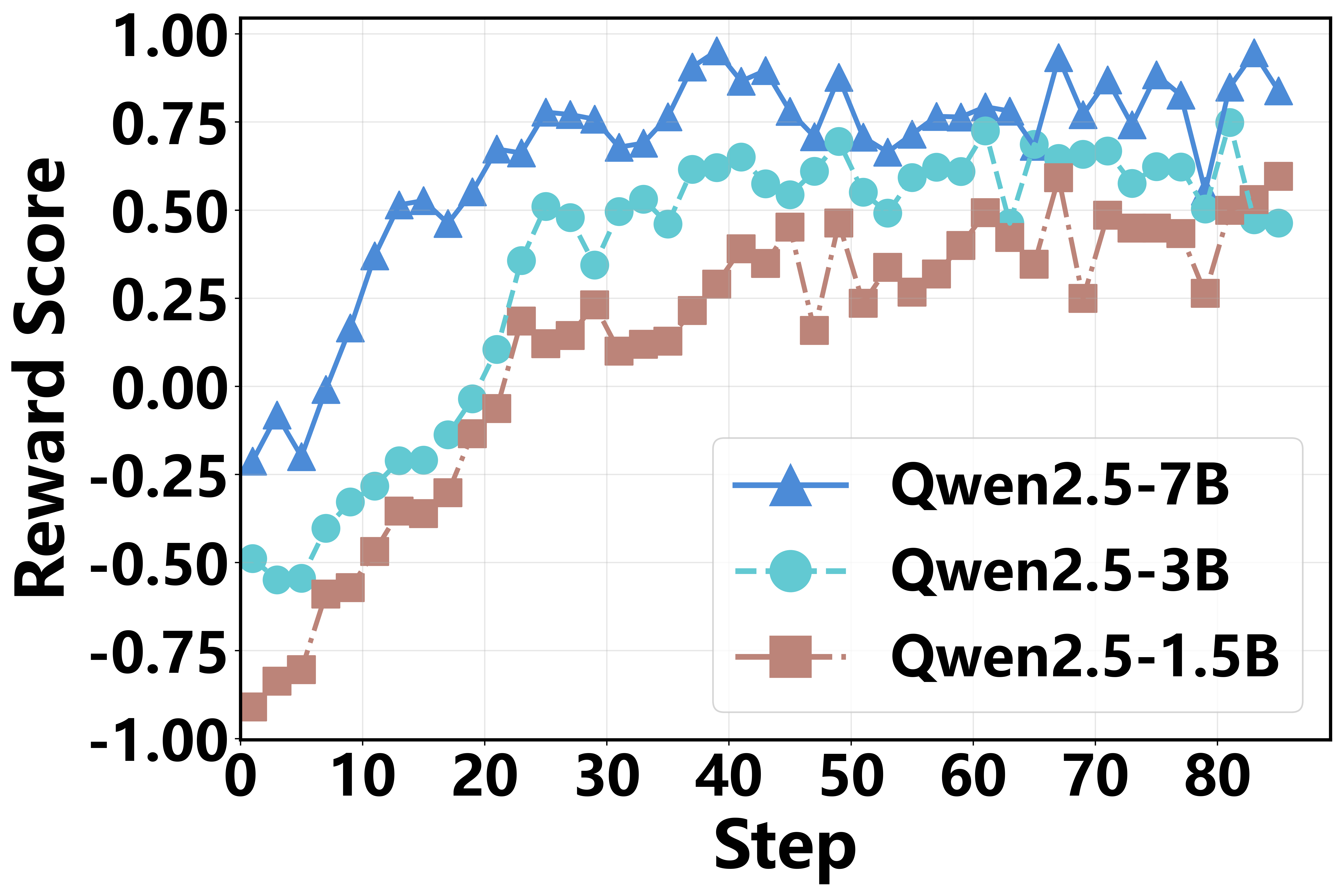}
        \caption{Model Scaling}
        \label{fig:compare_d}
    \end{subfigure}
    \vspace{-0.06in}
    \caption{(a) GRIP vs. GRPO: GRIP converges faster with slightly higher final performance. (b) RL vs. SFT+RL: SFT+RL achieves faster initial convergence and superior final performance. (c) Base vs. Instruct: Instruct model enables faster early reward growth, though base model achieves higher final reward. (d) Model Scaling: Larger models achieve higher final rewards, indicating stronger time-series reasoning and forecasting capability.}
    \vspace{-0.06in}
    \label{fig:compare}

\end{figure*}
\begin{table*}[t]\centering
\centering

\caption{Ablation study on Training Strategies, Reward Design, and Template Components.} 

\renewcommand{\arraystretch}{0.9}
\resizebox{0.85\textwidth}{!}{
\begin{tabular}{llcccccc}
\toprule
 & \multicolumn{1}{l}{\multirow{2}{*}{Method}} & \multicolumn{2}{c}{ETTh1} & \multicolumn{2}{c}{ETTm2} & \multicolumn{2}{c}{Wind} \\
\cmidrule(r{2pt}){3-4}
\cmidrule(l{2pt}r{2pt}){5-6}
\cmidrule(l{2pt}){7-8}
 & \multicolumn{1}{c}{}                  & MSE          & MAE        & MSE         & MAE         & MSE         & MAE        \\
\midrule

\cellcolor{lightgray!30}Full Model & \cellcolor{lightgray!30}\Ours                               & \cellcolor{lightgray!30}\textbf{5.8752}       & \cellcolor{lightgray!30}\textbf{1.2325}     & \cellcolor{lightgray!30}\textbf{5.6673}       & \cellcolor{lightgray!30}\textbf{1.3771}     & \cellcolor{lightgray!30}\textbf{1353.9381}    & \cellcolor{lightgray!30}\textbf{15.1095}   \\
\midrule
\multirow{2}{*}{\begin{tabular}[l]{@{}l@{}}Training\\ Strategies\end{tabular}} &\textit{w/o} SFT     & 6.3558       & 1.4278     & 6.3673       & 1.4850     & 1632.6491    & 16.7903   \\
 &\textit{w/o} RL                                & 13.2196      & 1.7820     & 12.5940      & 3.7759     & 3424.0485    & 28.1392   \\
\midrule
\multirow{5}{*}{Reward} &\textit{w/o} Length                   & 5.8781       & 1.2325     & 6.3210       & 1.4024     & 1358.5169    & 15.4612   \\
 &\textit{w/o} MSE                        & 10.0948      & 1.5614     & 9.7449       & 2.4865     & 2749.2582    & 21.0272   \\
 &\textit{w/o} Seasonal   Decomposition    & 6.0132       & 1.2403     & 6.0220       & 1.3859     & 1462.5454    & 15.8056   \\
 &\textit{w/o} Trend   Decomposition      & 7.4775       & 1.3429     & 6.6881       & 1.4523     & 1766.9750    & 16.1789   \\
 &\textit{w/o} Structural   Similarity    & 7.8558       & 1.4278     & 8.5940       & 1.7759     & 2316.0214    & 22.3412   \\
\midrule
\multirow{2}{*}{\begin{tabular}[l]{@{}l@{}}Template\\ Components\end{tabular}} &\textit{w/o} Timestamps    & 9.9146       & 1.5446     & 8.7454       & 2.4867     & 2816.0214    & 26.3412   \\
 &\textit{w/o} domain   context                  & 6.2286       & 1.3989     & 6.3517       & 1.5319     & 1639.4436    & 17.0182  \\
\bottomrule
\end{tabular}
}
\label{tab:ablation}

\end{table*}

\subsection{Main Results}

We implemented the \Ours framework on nine datasets. The comparison with baseline models is summarized in Table \ref{tab:main_results_mse}. A more comprehensive list of results for metrics such as MAE can be found in Appendix Table \ref{tab:main_results_mae}. Key observations are as follows:

(1) \textbf{Limitations of Classical Methods and LLM Baselines.} Traditional deep learning-based forecasting models, such as PatchTST, DLinear, and iTransformer, achieve reasonable performance but are limited by their one-step "fast thinking" paradigm, which primarily focuses on direct pattern fitting and struggles with complex temporal dependencies and high-level reasoning. LLM-based methods like TimeLLM perform better by leveraging the reasoning abilities of LLMs, especially for long-term and non-linear patterns. However, they still treat forecasting as a direct generation task without explicit step-by-step reasoning, making their predictions potentially inconsistent or logically flawed when facing complex temporal variations. Moreover, their reliance on pre-trained knowledge with minimal task-specific adaptation further limits the explainability and controllability of the forecasting process.

\vspace{0.02in}
(2) \textbf{Performance Improvement and Benefits of \Ours.} Our proposed \Ours follows a two-stage optimization framework. In the first stage, CoT-guided SFT enables the model to learn structured output formats and basic reasoning logic. In the second stage, we further enhance the model’s reasoning capabilities through GRIP with fine-grained reward mechanisms. These include logical consistency, temporal coherence, and multi-horizon accuracy, which iteratively refine the model's reasoning paths. Experimental results show this approach not only improves forecasting performance but also enhances generalization under zero-shot and out-of-distribution settings.

\begin{figure*}[h] \centering
    
    \makebox[0.32\textwidth]{ Exchange Dataset}
    \hspace{0.01\textwidth}
    \makebox[0.32\textwidth]{ AQShunyi Dataset}
    \hspace{0.02\textwidth}
    \makebox[0.31\textwidth]{ NASDAQ Dataset}
    \\
    \includegraphics[width=0.32\textwidth]{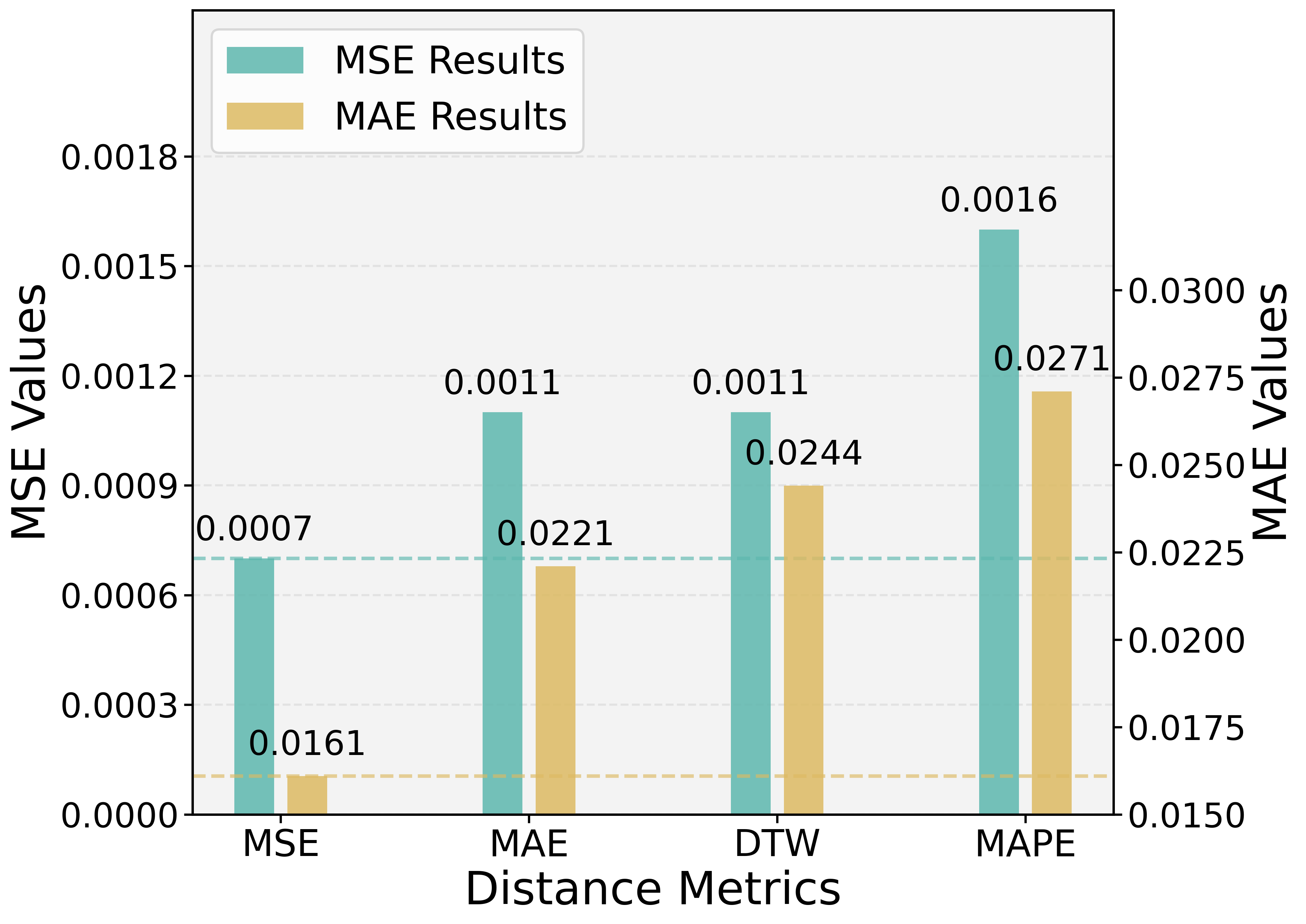}
     \hspace{0.01\textwidth}
    \includegraphics[width=0.32\textwidth]{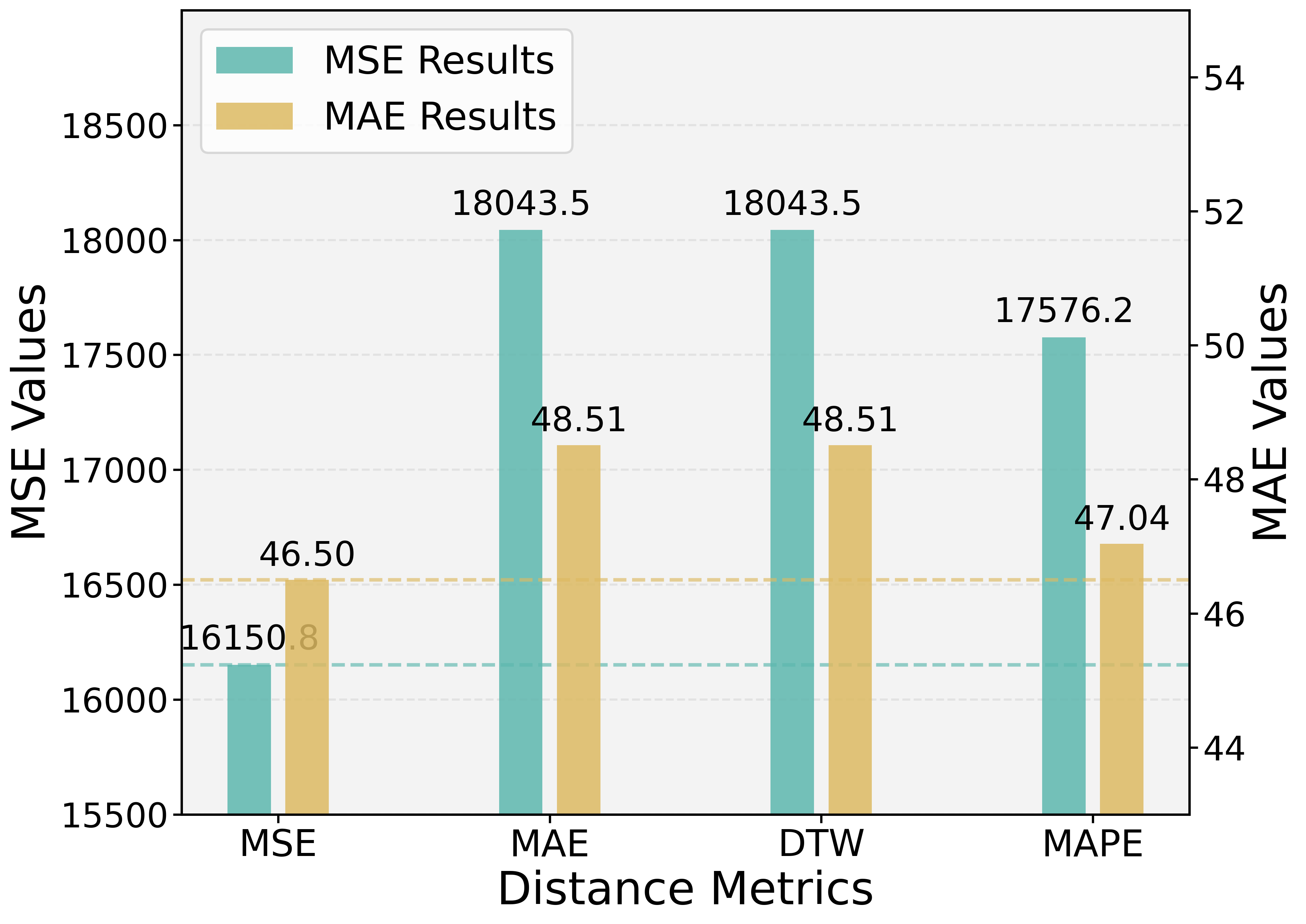}
    \hspace{0.01\textwidth}
    \includegraphics[width=0.32\textwidth]{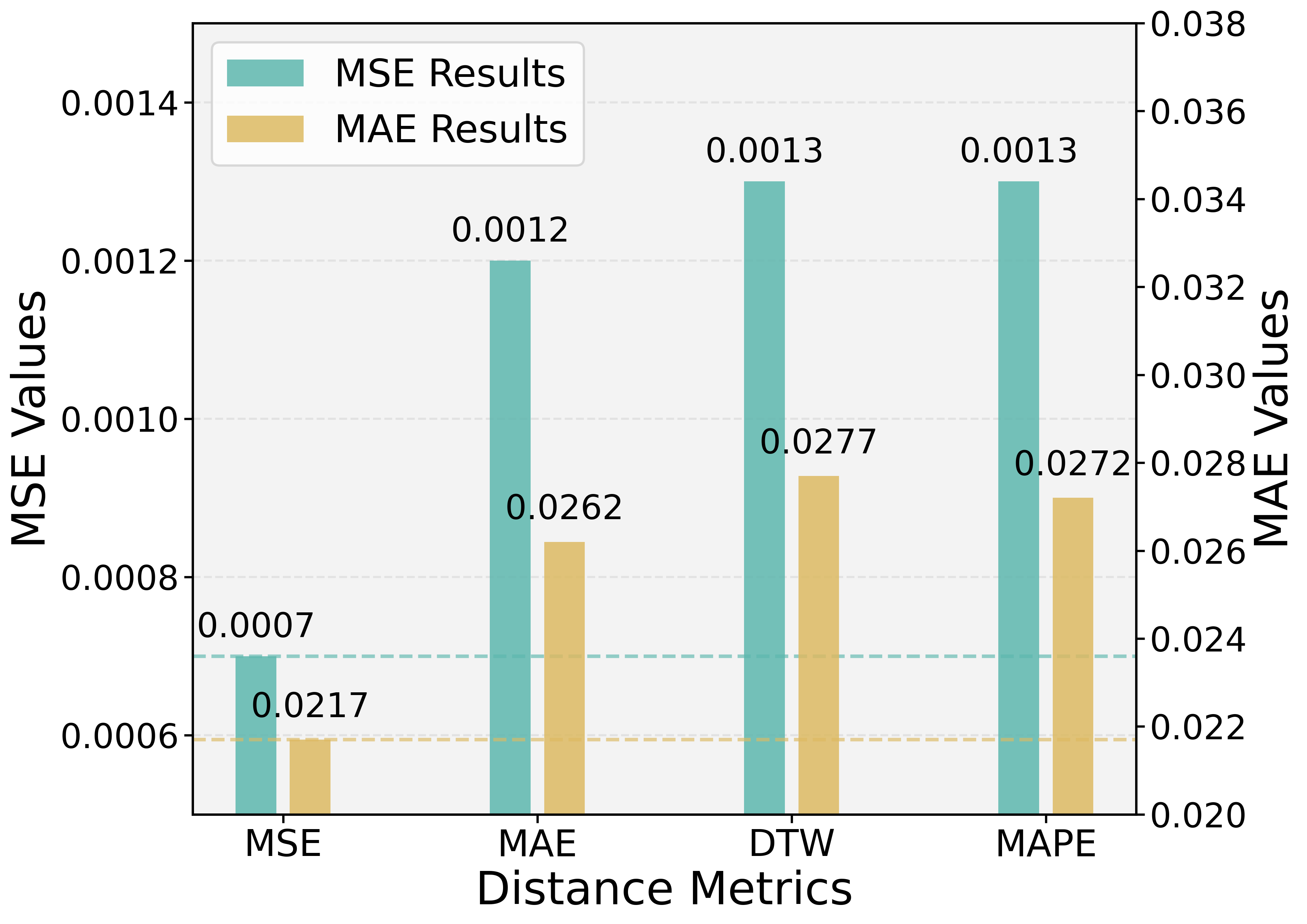}
    \vspace{-0.1in}
    \caption{Experimental results based on MSE, MAE, DTW, MAPE Distance Metrics.} 
    \label{fig:distance_metrics}
    \vspace{-0.1in}
\end{figure*}

\vspace{-0.02in}
\subsection{Ablation Study}


\paragraph{Impact of SFT and Reinforcement Learning.}



Next, we evaluate the necessity of CoT-based SFT by comparing two training strategies: (i) direct RL without SFT, and (ii) SFT followed by RL. As illustrated in Figure~\ref{fig:compare}(b), the model trained without SFT exhibits slower convergence and inferior performance in early training stages. In contrast, initializing RL with an SFT model accelerates learning and leads to better performance. This demonstrates that SFT provides a foundation for reasoning path generation, which is further refined through rule-augmented reinforcement learning.

Furthermore, we conducted an ablation study by removing RL (see Table \ref{tab:ablation}). The results show a degradation in TSF performance, with absolute performance drops on the ETT and Wind datasets, respectively, highlighting RL's crucial role in optimizing SFT-initialized reasoning paths and improving forecasting robustness. This indicates that while SFT establishes reasoning patterns, RL provides indispensable optimization through: (1) discovering higher-reward reasoning trajectories via exploration, and (2) suppressing plausible-yet-incorrect reasoning paths via reward shaping. Overall, RL is critical for achieving competitive performance.

\begin{table}[t]\centering
    \caption{Reasoning intervention study with No Think, Shuffled Think, and Corrupted Think. $\Delta$ denotes the relative MSE increase over Original Think.}
    \renewcommand{\arraystretch}{1.25}
    \vspace{-0.06in}
    \label{tab:reasoning_intervention}
    \resizebox{0.48\textwidth}{!}{
    \begin{tabular}{lccccccc}
        \toprule
        Dataset 
        & Original 
        & \multicolumn{2}{c}{No Think} 
        & \multicolumn{2}{c}{Shuffled Think} 
        & \multicolumn{2}{c}{Corrupted Think} \\
        \cmidrule(lr){2-2}
        \cmidrule(lr){3-4}
        \cmidrule(lr){5-6}
        \cmidrule(lr){7-8}
        & MSE$\downarrow$ 
        & MSE$\downarrow$ & $\Delta$ (\%) 
        & MSE$\downarrow$ & $\Delta$ (\%) 
        & MSE$\downarrow$ & $\Delta$ (\%) \\
        \midrule
        ETTh1 
        & \textbf{5.8752} 
        & 6.0629 & 3.19 
        & 6.5127 & 10.85 
        & 6.7031 & 14.09 \\
        ETTm2 
        & \textbf{5.6673} 
        & 5.9124 & 4.32 
        & 6.3186 & 11.49 
        & 6.5227 & 15.09 \\
        Wind 
        & \textbf{1353.9381 }
        & 1447.2865 & 6.89 
        & 1590.4173 & 17.47 
        & 1655.2947 & 22.26 \\
        \bottomrule
    \end{tabular}
    }
    \vspace{-0.2in}
\end{table}

\paragraph{Impact of Multi-objective Reward Design.}

We analyze the impact of each reward term in RL by training models with partial reward components. As shown in Table~\ref{tab:ablation}, removing any term degrades performance, indicating that all contribute to forecasting accuracy. The largest drops occur when MSE or Seasonal-Trend Decomposition rewards are removed, emphasizing the importance of point-wise precision and temporal structure. While Format and Length rewards have smaller effects on metrics, they ensure output consistency and training stability. Structural Similarity reward further enhances structural fidelity, especially for complex sequences. More experiments are in Section \ref{more_reward}.

\paragraph{Impact of Training Template Component.}

Finally, we assess how different elements of our structured prompts affect model behavior. We consider two main components: (i) explicit timestamp encoding, and (ii) contextual information such as seasonal period and task constraints. Table \ref{tab:ablation} shows that incorporating these components consistently improves both forecasting accuracy and generalization capability, especially under zero-shot and out-of-distribution scenarios. Models without timestamp information struggle to capture long-range dependencies, while those lacking contextual guidance often produce logically inconsistent outputs. 

\subsection{RL Optimizer Comparison}

We compare the performance of GRIP using two sampling strategies — Local Random Sampling, and Cluster-based Random Sampling — against GRPO, a policy optimization method in reasoning-based reinforcement learning. As illustrated in Figure~\ref{fig:compare}(a), the Cluster-based Random Sampling strategy achieves the highest performance, slightly outperforming GRPO. This is attributed to its ability to maintain diversity in trajectory selection by clustering samples based on reward values, which helps preserve potentially informative yet low-reward reasoning paths often ignored by greedy methods. In terms of convergence speed, Cluster-based Sampling also leads, followed by Local Sampling, and finally GRPO, which converges the slowest. Although local Sampling explores a larger search space, it tends to overfit high-reward trajectories early on, leading to poor generalization and suboptimal performance.

\begin{figure*}[htp]

    \centering
    \includegraphics[width=0.9\textwidth, keepaspectratio]{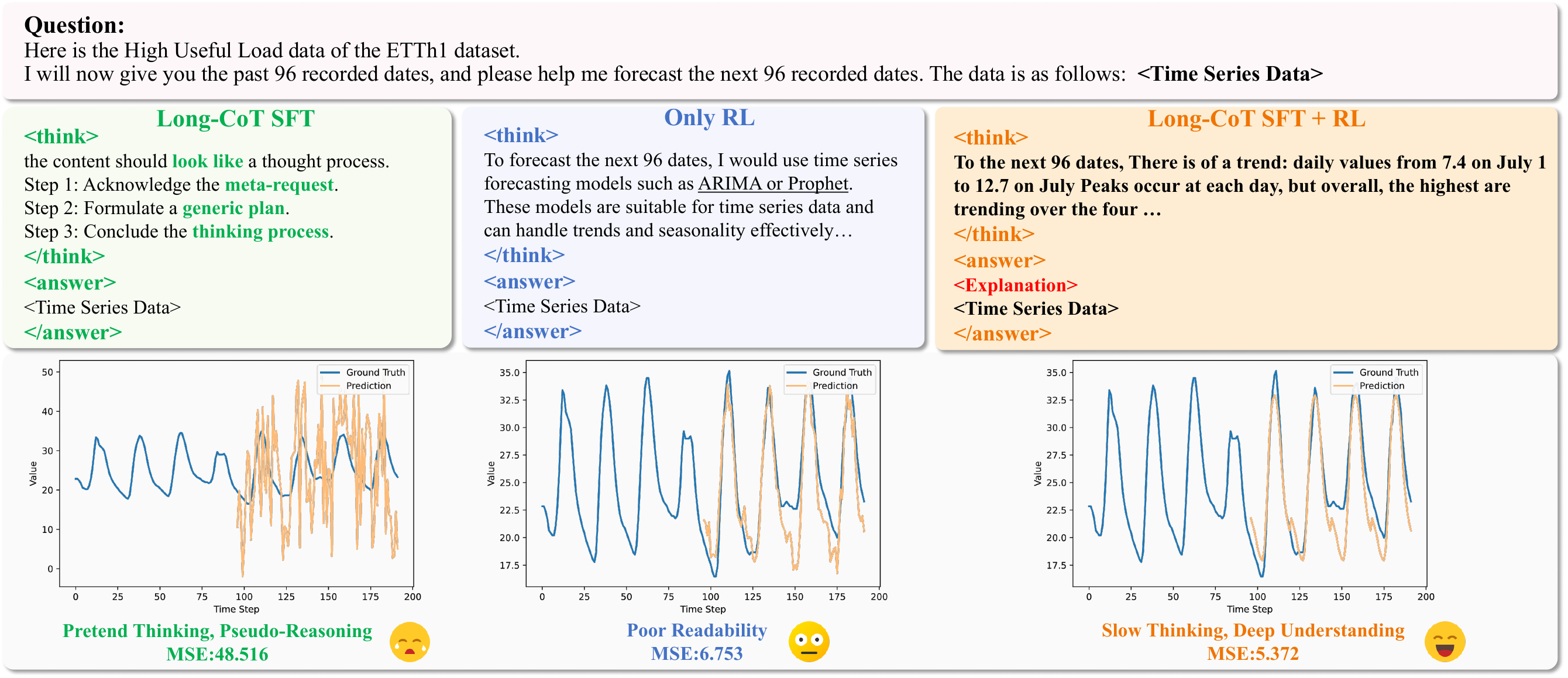} 
    \vspace{-0.12in}
    \caption{A reasoning case study of long-CoT SFT, RL, and Hybrid Methods on ETTh1 dataset.}
    \vspace{-0.12in}
    \label{fig:reasoning_case}

\end{figure*}

\vspace{-0.04in}
\subsection{Performance Comparison w.r.t Model Types}

We analyze the training dynamics of Time-R1 across two Qwen2.5 variants, Qwen2.5-7B-Base and Qwen2.5-7B-Instruct. Figure \ref{fig:compare}(c) shows the base model converges more slowly and starts from a lower performance level. However, it demonstrates stronger learning potential and eventually achieves slightly better results. This suggests that while instruction tuning accelerates early learning in time series reasoning, iterative RL-based optimization enables the base model to reach marginally superior performance.

\vspace{-0.04in}
\subsection{Impact of Accuracy Reward Metrics}
\label{more_reward}
To examine the choice of accuracy reward in GRIP, we compare four distance metrics, including MSE, MAE, DTW, and MAPE, on the Exchange, AQShunyi, and NASDAQ datasets, as shown in Figure~\ref{fig:distance_metrics}. Overall, MSE yields the best performance across datasets, with MAE ranking second, while DTW and MAPE bring limited gains. This suggests that an effective accuracy reward should provide a stable, discriminative signal aligned with the forecasting objective. Compared with MAE, MSE more strongly penalizes large deviations and better guides the model to reduce severe multi-step forecasting errors. In contrast, DTW focuses on shape alignment but may weaken point-wise accuracy, and MAPE can be unstable when target values are close to zero. Based on these observations, we adopt MSE as the primary accuracy reward in our framework.

\subsection{Performance Comparison w.r.t Model Sizes} 

To evaluate the scaling behavior of \Ours, we conduct experiments using models with 1.5B, 3B, and 7B parameters on TSF tasks. As shown in Figure~\ref{fig:compare}(d), forecasting performance consistently improves with increasing model size. The 1.5B model achieves reasonable results on simple datasets but struggles with complex temporal patterns. In contrast, the 7B model achieves the best overall performance, particularly in capturing long-term dependencies and handling out-of-distribution scenarios. These results indicate that larger models enhance temporal reasoning capabilities and that \Ours effectively benefits from stronger backbone models.

\subsection{Reasoning-Format Intervention Study}
To examine whether reasoning contributes to forecasting performance, we conduct a reasoning-format intervention study. For the No Think variant, we modify the reward so that the model is no longer encouraged to generate reasoning, but directly outputs forecasts in \texttt{<answer>...</answer>}.

To test whether reasoning content affects prediction, we construct two reasoning-perturbation settings at inference time. In Shuffled Think, we collect the \texttt{<think>} outputs generated by Time-R1 on the test set and replace each sample's reasoning with a \texttt{<think>} from another sample in the dataset, while keeping the historical input unchanged. This creates fluent but input-mismatched reasoning. In Corrupted Think, we rewrite the reasoning by reversing key temporal conclusions, such as changing increasing trends to decreasing trends, weakening or removing seasonality, and reversing volatility or level-shift descriptions, while preserving format and style. As shown in Table~\ref{tab:reasoning_intervention}, Original Think achieves the lowest MSE across ETTh1, ETTm2, and Wind, while removing, mismatching, or corrupting reasoning degrades performance. These results suggest that reasoning encouraged during training is associated with forecasting accuracy, and that its content is not merely a superficial explanation but can influence the final prediction.

\subsection{Visualization of the Reasoning Process}

As shown in Figure \ref{fig:reasoning_case}, our case study highlights key differences among training paradigms. SFT enables imitation of reasoning patterns but often results in superficial replication, leading to flawed logic and suboptimal performance. Pure RL achieves reasonable accuracy but generates reasoning traces with poor readability. In contrast, the SFT+RL paradigm not only teaches extended reasoning effectively but, through its reinforcement phase, also improves prediction accuracy while helping the model identify which reasoning components most contribute to performance gains.

\vspace{0.08in}
\section{Conclusion}


In this work, we proposed \Ours, a generative time series forecasting framework that enables LLMs to perform deliberate step-by-step reasoning for improved prediction across diverse real-world forecasting datasets. We introduced time series reasoning by training LLMs to adopt a slow-thinking paradigm, generating clear and explainable intermediate reasoning steps before producing final forecasts, which achieves competitive TSF performance. 
Experiments demonstrate that model scaling enables substantial and consistent improvements in time series reasoning quality, while verified-reward RL yields stronger generalization to out-of-domain tasks. 
We hope this work paves the way for future research in structured temporal reasoning.




\begin{acks}
This work was supported by grants from the National Natural Science Foundation of China (62502486), the Fundamental Research Funds for the Central Universities of China, USTC Kunpeng-Ascend Scientific and Education Innovation Excellence Center.
\end{acks}

\clearpage

\appendix
\clearpage
\section*{Appendix}

{
\renewcommand{\arraystretch}{1.13}
\begin{table*}[!b]
    \centering
    \vspace{-0.1in}
    \caption{Comprehensive performance comparison of \Ours and various baseline models with best values in bold and second-best underlined. MAE $\downarrow$ is specifically used as the evaluation metric. DeepSeek-R1 denotes zero-shot results using our reasoning template. Overall, \Ours achieves competitive or best performance on most datasets.}
    \label{tab:main_results_mae}
    \vspace{-0.1in}
    \resizebox{0.98\textwidth}{!}{
    \large
    \begin{tabular}{ll|ccccccccc}
        \toprule
         & \textbf{Methods}    & \textbf{ETTh1}  & \textbf{ETTh2}   & \textbf{ETTm1}   & \textbf{ETTm2}  & \textbf{Exchange} & \textbf{AQWan} & \textbf{AQShunyi} & \textbf{Wind} & \textbf{NASDAQ} \\
        \hline
        \multirow{11}{*}{Traditional} & AutoFormer & 1.5945 & 2.3246 & 2.0989 & 1.7696 & 0.0230 & 48.4358 & 51.4833 & 19.6086 & 0.0229 \\
        & PatchTST & 1.6376 & 2.0076 & 1.9446 & \underline{1.3973} & 0.0200 & \underline{39.5089} & 42.5427 & 20.4654 & \textbf{0.0213}\\
        & DLinear & 1.5116 & 2.0855 & 1.7815 & 1.9136 & 0.0256 & 51.7896 & 50.0134 & 18.6036 & 0.0217 \\
        & iTransformer & 1.5126 & 1.9295 & 1.6609 & 1.4598 & 0.0204 & 40.3070 & 42.5116 & 18.1864 & 0.0237 \\
        & TimeXer & 1.5596 & 2.0895 & 1.7785 & 1.4419 & 0.0195 & 41.8280 & 41.4929 & 18.4876 & 0.0233 \\
        & TimeMixer & \underline{1.4018} & 2.0456 & \underline{1.7103} & 1.3987 & \underline{0.0182} & 40.6789 & \textbf{40.3452} & 18.8801 & 0.0219 \\
        & WPMixer & 1.4321 & 2.0013 & 1.7502 & 1.4205 & 0.0188 & 41.2347 & \underline{40.9811} & \underline{18.1034} & 0.0225 \\
        & Moment & 1.4789 & 1.9912 & 1.7654 & 1.4456 & 0.0188 & 40.5432 & 42.1234 & 18.6789 & 0.0221 \\
        & TimesFM-2.5 & 1.4922 & 1.7195 & 2.7197 & 1.7070 & 0.0193 & 40.0290 & 43.6785 & 18.3456 & \underline{0.0215} \\
        & Chronos-2 & 1.5108 & 2.0533 & 2.5785 & 1.6766 & 0.0196 & 40.3737 & 42.3991 & 20.1270 & 0.0268 \\
        \hline
        \multirow{4}{*}{LLM-based} & CrossTimeNet & 1.5492 & 2.1731 & 1.9863 & 1.5281 & 0.0217 & 43.6267 & 46.0941 & 20.6205 & 0.0252 \\        
        & GPT4TS & 1.4299 & \underline{1.9134} & 1.9635 & 1.4943 & 0.0202 & 40.2815 & 42.3969 & 18.4919 & 0.0256 \\
        & TimeLLM & 1.4286 & 1.9143 & 1.9626 & 1.4926 & 0.0201 & 40.2746 & 42.3926 & 18.4890 & 0.0255 \\
        & DeepSeek-R1 & 1.4316 & 2.0165 & 1.8876 & 1.6289 & 0.0209 & 64.9627 & 57.6412 & 30.6469 & 0.0231 \\
        \hline
        \cellcolor{lightgray!30}Ours & \cellcolor{lightgray!30}Time-R1 & \cellcolor{lightgray!30}\textbf{1.2325} & \cellcolor{lightgray!30}\textbf{1.7192} & \cellcolor{lightgray!30}\textbf{1.6510} & \cellcolor{lightgray!30}\textbf{1.3771} & \cellcolor{lightgray!30}\textbf{0.0161} & \cellcolor{lightgray!30}\textbf{39.1846} & \cellcolor{lightgray!30}46.5032 & \cellcolor{lightgray!30}\textbf{15.1095} & \cellcolor{lightgray!30}0.0217 \\
        \bottomrule
    \end{tabular}
    }
    \vspace{-0.1in}
\end{table*}
}

\begin{figure*}[!b]
    \centering
    \includegraphics[width=\textwidth]{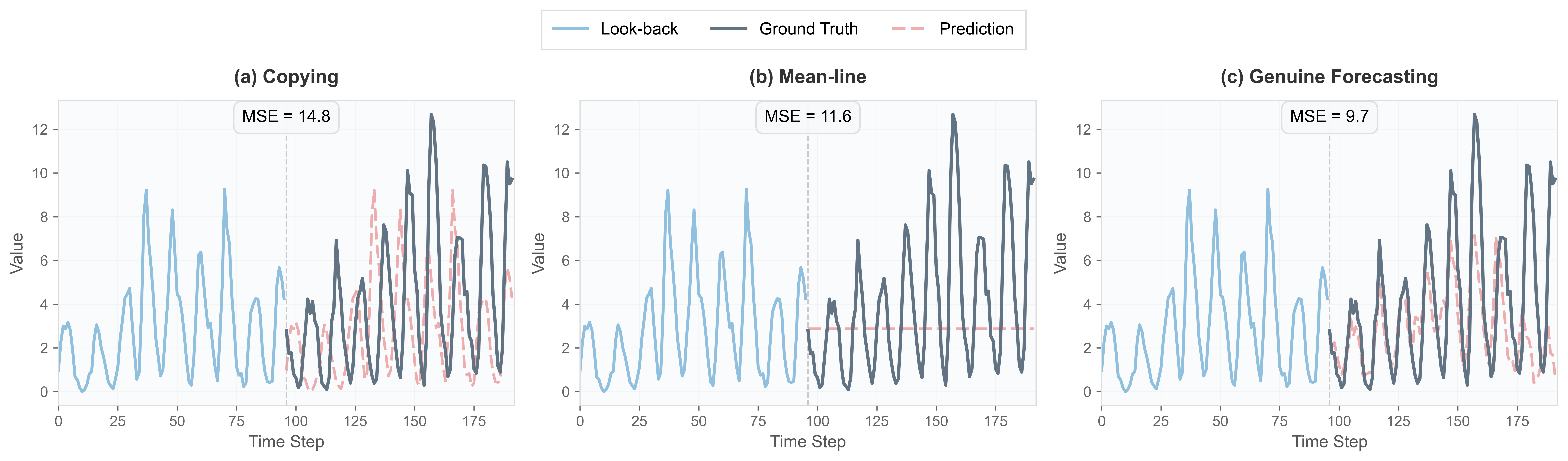}
    \vspace{-0.3in}
    \caption{Three representative rollout modes for 96$\rightarrow$96 forecasting: 
    (a) copying the look-back window, 
    (b) mean-line output, and 
    (c) genuine forecasting. 
    Although these trajectories are qualitatively different, their MSE values can still be close enough to weaken the learning signal, especially in non-binary forecasting tasks.}

    \label{fig:forecasting_comparison}
\end{figure*}

\section{GRPO vs. GRIP: Binary-Reward Reasoning vs. Non-Binary Time-Series Forecasting} \label{section:po.vs.ip}
GRPO is well suited to reasoning tasks with \emph{verified} outcomes, such as mathematical problem solving, where rewards are nearly binary and group-relative advantages provide sharp learning signals by consistently reinforcing correct trajectories. Time-series forecasting, however, is inherently non-binary and graded under metrics such as MSE or MAE, where predictions often vary continuously in quality rather than being clearly correct or incorrect. In this setting, the reward differences among sampled completions may not reliably distinguish genuinely predictive trajectories from merely less-bad ones. More critically, when an entire sampled group consists of mediocre forecasts, the group-relative normalization in GRPO can still assign positive advantages to the relatively better samples, thereby producing misleading gradient signals that reinforce suboptimal behaviors. This may encourage ``safe'' but uninformative patterns, such as copying the look-back window or producing a flat mean-line forecast, especially when high-quality predictive trajectories are sparse. To better match this continuous reward landscape, GRIP introduces non-uniform sampling and adaptive weighting: it first explores a larger candidate pool of $k\!\cdot\!G$ trajectories, then selects $G$ trajectories through reward-weighted or diversity-preserving sampling, and finally amplifies the gradient contribution of high-reward completions rather than treating all samples equally. As illustrated in Figure \ref{fig:forecasting_comparison}, this design increases the probability of selecting and reinforcing genuine forecasting trajectories that capture trend, seasonality, and local variations.

\section{Main Results}
We present the main results using the MAE evaluation metric here in Table \ref{tab:main_results_mae}. Compared to both traditional methods and LLM-based approaches, \Ours achieves competitive improvements.

\clearpage
\section*{GenAI Usage Disclosure}
GenAI Usage Disclosure. DeepSeek-R1 was used to generate candidate forecasts and label-conditioned synthetic reasoning trajectories for constructing the supervised fine-tuning data described in Section 3.4. Generative AI tools were additionally used for language polishing. The scientific contributions, methodological design, experimental findings, and interpretations were developed independently by the authors throughout the research process.

\bibliographystyle{ACM-Reference-Format}
\bibliography{ref}


\end{document}